\documentclass[conference]{IEEEtran} 
\pdfoutput=1
\IEEEoverridecommandlockouts

\usepackage{cite}
\usepackage{amsmath,amssymb,amsfonts}
\usepackage{algorithmic}
\usepackage{graphicx}
\usepackage{textcomp}
\usepackage{xcolor}

\usepackage{graphicx}
\usepackage{subcaption}
\usepackage{multirow}
\usepackage{fancyhdr}
\usepackage{comment}
\usepackage{subfiles}

\usepackage[breaklinks]{hyperref}
\IEEEoverridecommandlockouts

\begin{document}

\title{MDSF: Context-Aware Multi-Dimensional Data Storytelling Framework based on Large language Model}

\author{
    \IEEEauthorblockN{
        1\textsuperscript{st} Chengze Zhang\textsuperscript{*},
        2\textsuperscript{nd} Changshan Li
        3\textsuperscript{rd} Shiyang Gao,
    }
    \IEEEauthorblockA{
        \textit{MITC, Xiaomi Corp.}\\
        Beijing, China \\
        zhangchengze, lichangshan, gaoshiyang@xiaomi.com
    }
    \thanks{* Chengze Zhang is the corresponding author.}
}

\maketitle

\begin{abstract}

The exponential growth of data and advancements in big data technologies have created a demand for more efficient and automated approaches to data analysis and storytelling.
However, automated data analysis systems still face challenges in leveraging large language models (LLMs) for data insight discovery, augmented analysis, and data storytelling. 
This paper introduces the Multidimensional Data Storytelling Framework (MDSF) based on large language models for automated insight generation and context-aware storytelling. 
The framework incorporates advanced preprocessing techniques, augmented analysis algorithms, and a unique scoring mechanism to identify and prioritize actionable insights. 
The use of fine-tuned LLMs enhances contextual understanding and generates narratives with minimal manual intervention. 
The architecture also includes an agent-based mechanism for real-time storytelling continuation control.
Key findings reveal that MDSF outperforms existing methods across various datasets in terms of insight ranking accuracy, descriptive quality, and narrative coherence. 
The experimental evaluation demonstrates MDSF's ability to automate complex analytical tasks, reduce interpretive biases, and improve user satisfaction. 
User studies further underscore its practical utility in enhancing content structure, conclusion extraction, and richness of detail.

\end{abstract}


\section{Introduction}



The explosive growth of data scale and the rapid development of big data technology have driven progress across various fields, but have also introduced the challenge of analyzing massive amounts of data. Efficient data analysis and precise data interpretation have thus become crucial tasks for data scientists and analysts.
The traditional data analysis pipeline struggles to achieve comprehensive and accurate insights in big data scenarios. Intelligent data storytelling, as an important method of data analysis, is poised to reconstruct the data analysis process.
This method is based on LLM agents and comprises various capabilities such as data visualization, storytelling techniques, and contextual analysis, bringing a new paradigm to the field of data analysis.
According to Gartner, by 2025, intelligent data storytelling will become the primary mode of analytic consumption, with 75\% of data stories predicted to be automatically generated using augmented intelligence and machine learning, rather than by data analysts.

\begin{figure}
    \centering
    \includegraphics[width=1\linewidth, trim=6cm 6.7cm 6.5cm 2cm, clip]{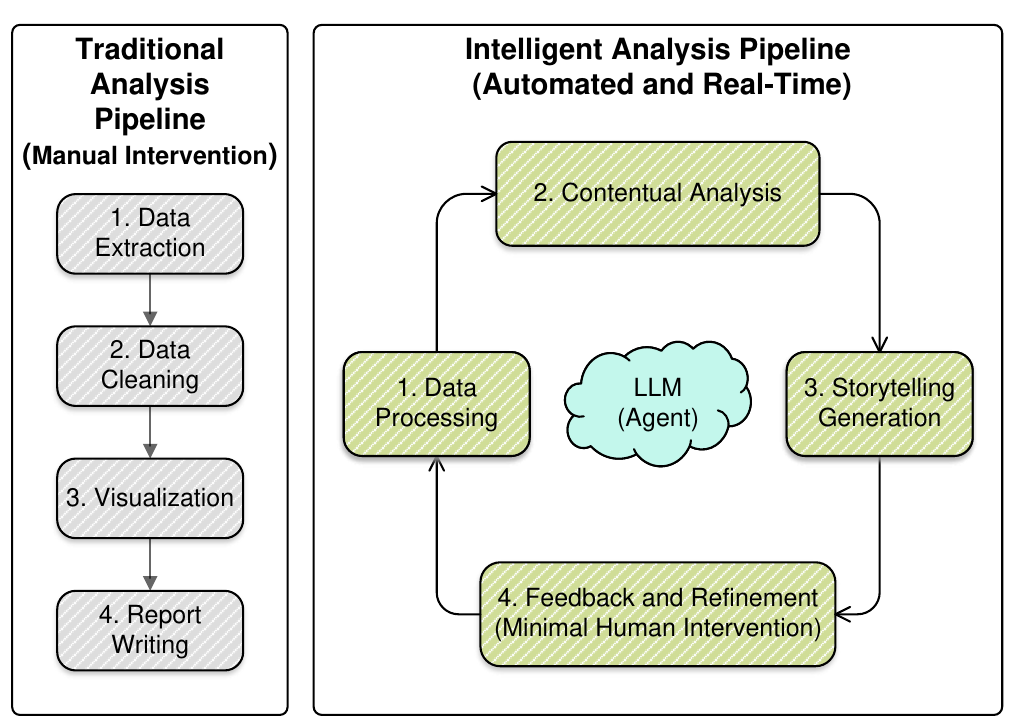}
    \caption{Comparison of traditional and intelligent data analysis pipelines. }
    \label{fig:enter-label}
\end{figure}


Intelligent Data Storytelling (IDS) holds significant advantages in big data environments by automating labor-intensive steps within the traditional data analysis pipeline, thereby establishing a data-driven and efficient analysis process.
With the rapid advancement of large model technologies, the generalization and comprehension capabilities of these models present new opportunities for data storytelling. IDS leverages the extensive prior knowledge embedded in large models to automatically generate data reports during user interactions, significantly enhancing efficiency and reducing interpretative biases.
For instance, in corporate sales analysis, the traditional analysis pipeline typically involves multiple stages: data extraction, cleaning, visualization creation, and report writing, each heavily dependent on the analyst's involvement, resulting in a lengthy process.
In contrast, IDS automates data cleaning and dynamically generates narratives, presenting data insights directly in visual and natural language formats. It can also capture anomalies in real-time and provide explanations. This intelligent process eliminates the need for sequential tool switching, greatly optimizing the speed and quality of data interpretation.

\begin{table}[htp]
    \centering
    \caption{Key comparison factors between traditional and intelligent pipelines.}
    \label{tab:comparison}
    \begin{tabular}{lcc}
        \hline
        \textbf{Factor}       & \textbf{Traditional} & \textbf{Intelligent} \\ \hline
        Efficiency            & Low                  & High                 \\ \hline
        Consistency           & Prone to Bias        & Reliable             \\ \hline
        Scalability           & Limited              & Scalable             \\ \hline
    \end{tabular}
\end{table}




Despite the growing importance of data storytelling, current intelligent data storytelling still faces numerous challenges. First, most existing methods are typically limited to single-dimensional data analysis, making it difficult to effectively integrate and present complex multi-dimensional data. Second, generating accurate and engaging narratives remains a time-consuming process that requires specialized skills, which restricts the widespread application of data storytelling across various fields. Additionally, existing tools often lack sufficient flexibility and adaptability to meet the diverse needs of different industries and audiences.



To address these challenges, we propose the Multi-dimensional Data Storytelling Framework (MDSF). MDSF is based on Large Language Models (LLMs) and aims to provide a comprehensive, flexible, and intelligent solution to overcome the limitations of current data storytelling methods. It can simultaneously process and integrate multi-dimensional data, including numerical, categorical, time-series, and spatial data, thereby generating more comprehensive and in-depth data narratives. By leveraging the powerful natural language processing and generation capabilities of LLMs, MDSF can automatically create coherent and engaging stories, significantly reducing the need for manual intervention and increasing efficiency.




Overall, the introduction of MDSF aims to advance the technology of data storytelling, providing more powerful and intelligent tools to various fields for more effectively extracting data value, promoting information dissemination, and facilitating decision-making. By addressing the limitations of existing methods, MDSF is expected to bring revolutionary changes to data analysis and communication, making data storytelling a more prevalent, efficient, and insightful practice. The major contributions of this paper are as follows: 

\begin{enumerate}
    \item Development of the Multi-dimensional Data Storytelling Framework (MDSF): We propose an innovative framework based on Large Language Models (LLMs) for multi-dimensional data storytelling. This framework can handle complex datasets and generate easily understandable narrative content. 
    \item Integration of LLMs into the Data Storytelling Process: We successfully integrate Large Language Models into the data storytelling process, ensuring that the generated narratives are more coherent and insightful. This approach not only improves the quality of the narratives but also significantly reduces the need for manual intervention.
    \item Empirical Validation of MDSF's Effectiveness and Advantages: We demonstrate the effectiveness and advantages of MDSF through a series of experiments and case studies. The results indicate that this framework performs excellently in data storytelling tasks across different fields, significantly enhancing the efficiency of data analysis and decision-making.
\end{enumerate}

\section{Related Works}
In recent years, with the continuous growth of data scale and the increasing complexity of analysis, both academia and industry have been deeply researching efficient data analysis methods. Relevant work can be summarized into three main directions: automated data analysis, the application of large language models in data insights, and automated report generation based on agents. These studies provide the theoretical foundation and practical reference for the proposed Multi-dimensional Data Storytelling Framework (MDSF).

\subsection{Automated Data Analysis \& Augmented Analysis}
Automated data analysis and augmented analytics have gained significant attention due to their ability to streamline data-driven decision-making processes. Islam et al. \cite{1} introduced DataNarrative, an innovative framework combining visualization and textual storytelling to enhance automated insights. Similarly, CoInsight by Li et al. \cite{6} focuses on hierarchical table storytelling, emphasizing the utility of visual storytelling techniques to connect data insights. 

Xie et al. \cite{4} developed HAIChart, which pairs human and AI capabilities for visualization, bridging the gap between raw data and actionable insights. Additionally, Shao et al. \cite{8} explored the impact of storytelling in data visualization, finding that it improves both the efficiency and effectiveness of insights comprehension. These works highlight the evolution of data analysis techniques, combining automation with human-centric designs to improve the interpretability and usability of data-driven narratives.

\subsection{Large Language Models in Data Insight}
Large Language Models (LLMs) have been pivotal in enhancing data insight capabilities. Zhu et al. \cite{5} investigated the statistical acumen of LLMs, providing a benchmark for their application in complex data analysis scenarios. Sui et al. \cite{12} examined LLMs' understanding of structured table data, offering insights into their strengths and limitations in handling tabular information.

He et al. \cite{13} conducted a comprehensive survey on leveraging LLMs for narrative visualization, emphasizing their role in crafting coherent and engaging data stories. Furthermore, InsightPilot by Ding et al. \cite{17} exemplifies an LLM-powered system for automated data exploration, showcasing their potential in streamlining exploratory data analysis tasks. These studies underscore the transformative impact of LLMs on data insight generation and analysis, enabling more nuanced and scalable approaches to understanding complex datasets.

\subsection{Agent-based Automated Reporting}
Agent-based systems have emerged as a promising paradigm for automated reporting. DS-Agent, proposed by Guo et al. \cite{7}, integrates large language models with case-based reasoning to automate data science workflows, significantly reducing manual effort. Shen et al. \cite{14} extended this concept with an LLM-based multi-agent system for creating animated data videos, pushing the boundaries of automated narrative generation.

Additionally, Singha et al. \cite{9} explored semantically aligned question and code generation for insight generation, demonstrating the efficacy of agents in automating complex analytical processes. The Melody platform by Renda et al. \cite{27} further exemplifies agent-based storytelling, combining linked open data visualization with curated narratives. These advancements highlight the growing role of agent-based frameworks in delivering dynamic and contextually relevant reports.

\section{Preliminaries}

\subsection{Data Model}


The data model \( D \) can be abstracted as an \( N \times M \) data table, where \( N \) represents the number of rows and \( M \) represents the number of dimensions. The dimensional attributes include time dimensions, categorical dimensions, and numerical dimensions, corresponding to time series data, categorical data, and numerical data, respectively.

A \textbf{breakdown} dimension is a dimension on which enumeration operations are performed on the data model \( D \). When a breakdown is specified, the indicators in the data model can be calculated by executing aggregation functions.
An \textbf{indicator} is a function attribute that aggregates data. For example, an indicator measuring overall revenue is calculated by summing the individual revenue data points.
A \textbf{filter} is a conditional selection applied to the data model \( D \). For example, a filter can restrict the data domain to a specific time range or select a particular attribute with a specific value.

For instance, in a sales data model, the breakdown dimension could be the region, the indicator could be the sales amount, and the filter could be the time range. It is important to note that in practical business systems, the time dimension and time filters are often specified separately to optimize performance.

\subsection{Data Insight}

A data insight is a specific pattern identified within a data model, discovered through our predefined types of insights and identification methods. An insight can be represented by the following components:

\begin{equation}
    Insight := \left\langle
    \begin{aligned}
        &breakdowns, indicators, type,\\
        &datamodel, details, score
    \end{aligned}
    \right\rangle
\end{equation}


Insights are designed to encompass various types to describe specific patterns in the data model from different perspectives. These include describing distribution characteristics of fundamental attributes, year-over-year and month-over-month comparisons, detecting anomalies, periodicity, and trends in time series data, as well as more complex analyses such as root cause analysis and correlation breakdowns. Additionally, each insight includes a specific score, which quantifies the value of the insight. This score is crucial for large language models to understand and assess the significance of the insights.

\subsection{Data Story}

A data story is composed of multiple insights. By leveraging large language models, descriptions of various insights are generated and integrated into coherent paragraphs, along with visual components, to form a comprehensive data report. 

The report includes a summary, detailed descriptions of the findings, a list of the source insights, and visual components such as charts and tables to present the data more intuitively.

\section{The MDSF}

\subsection{Insight Discovery}

\begin{figure*}[htp]
    \includegraphics[width=1\textwidth, trim=10cm 8.2cm 4cm 8.9cm, clip]{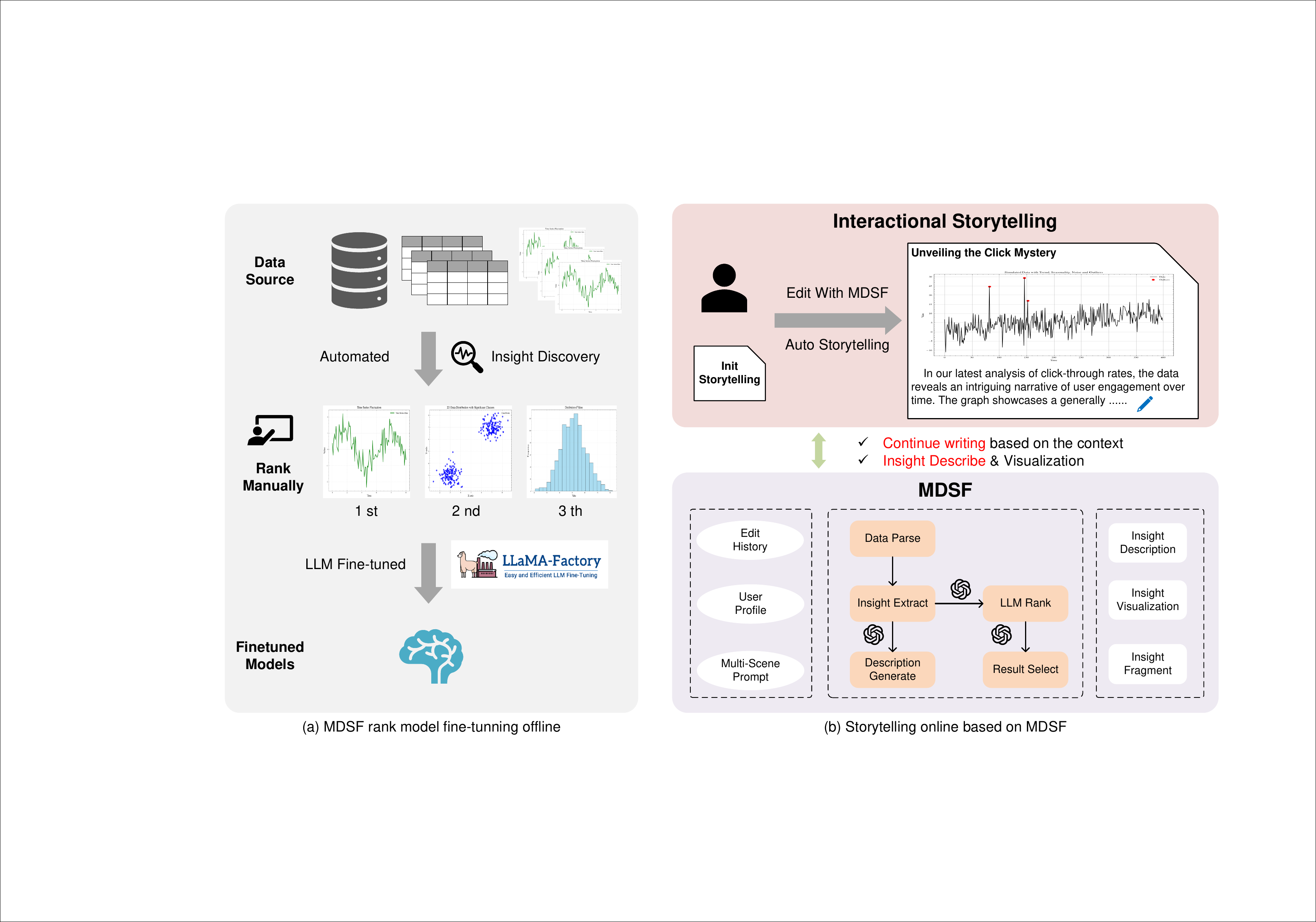}
    \caption{An Overview of MDSF}
    \label{fig:framework}
\end{figure*}


We have designed a Multidimensional Data Storytelling Framework (MDSF) to automatically generate data insight text and provide real-time report generation capabilities based on context when users edit reports.

\subsubsection{Data Parse \& Preprocess}



Data identification and preprocessing are the primary steps in the Multidimensional Data Storytelling Framework (MDSF). The system automatically extracts data from various sources (such as databases and file systems) and performs necessary preprocessing operations, including data cleaning, format conversion, and missing value imputation. This process ensures the accuracy and consistency of subsequent analyses. By utilizing automated tools, we significantly reduce manual intervention and increase data processing efficiency.

Firstly, we clean the underlying multidimensional data and enumerate different subspaces based on filtering operations and dimension selection. Each subspace can derive multiple peer subspaces according to different filtering conditions. After grouping and aggregation, we obtain a finer granularity of the data domain. Data preprocessing involves subspace enumeration, pre-computation of data aggregation, and indicator trimming operations. 

Trimming primarily includes: dimension trimming based on correlation, and data filtering based on the null value rate and coverage rate of the indicators.

\subsubsection{Augmented Analysis Algorithm}


In the process of discovering insights, we employ various augmented analysis algorithms. For time series data, we use algorithms such as Prophet and SR-CNN. For non-time series data, we utilize a combination of the 3-sigma and iForest algorithms, as well as time series forecasting methods

\subsubsection{Insight Score \& Rank}


To evaluate and rank insights, we design a multi-angle scoring mechanism and employ a Top-K algorithm for ranking.
The scoring mechanism includes the following aspects:

\paragraph{Importance}


Evaluate the importance of the subspaces used to generate insights for the complete dataset. The greater the significance of a subspace within the collection, the higher its importance score. Based on practical data analysis experience, it is observed that long-tail distribution is a common phenomenon in datasets. The benefits of focusing on tail data are much less than those of focusing on head data. Therefore, the distribution characteristics are used here to quantify the importance of the data.

\begin{equation}
      Imp = \frac{ |x_i-\overline{x}| }{ \sum_{i=0}^{n}{x_i} } * \frac{AscRN_x}{n+1}
\end{equation}

\paragraph{Significance}


The value score is a measure of the Significance of the insight.
According to the evaluation of the insight judgment function, the degree of matching with the type, specifically, the significant degree of anomaly, the degree of fitting with the distribution, etc., varies with the different insight types.

\paragraph{Surprise}



Surprise score is used to evaluate the degree to which users are surprised by the insight results.
The degree of surprise mainly includes two aspects: on the one hand, whether the result is beyond the user's expectation of satisfactory results; On the other hand, whether the insight results can be easily found by the user drill-down.
The purpose or expectation of an insight algorithm is to find insights that are difficult for analysts to find.
In this scenario, the Insight surprise score that is easy to be found is lower than that of the Insight surprise score that is difficult to be found.

\begin{equation}
    \begin{aligned}
        &srp_{x} = JS(P, Q) \\
        &\phantom{Srp_{x}} = \frac{1}{2} \left( \sum_{i} p_i \log \frac{2p_i}{p_i + q_i} + \sum_{i} q_i \log \frac{2q_i}{p_i + q_i} \right) \\
        &srp = \frac{\sum_{x \in X} srp_{x}}{\text{len}(X)}
    \end{aligned}
\end{equation}

$P$ is the data distribution of the sibling subspace, $Q$ is the data distribution of the native space, $x$ is the enumerated value of the subspace, and $X$ is the enumerated dimension of the subspace.

\paragraph{Fatigue}



Fatigue refers to the user's tolerance for similar or identical insights. 
When the recommended insights in the recommendation pool are too similar and do not align with user preferences, user fatigue can occur, negatively impacting the effectiveness of the insights. 
Therefore, fatigue is used to score and rank insights according to specific rules. 
The design of fatigue consists of two components:
\begin{itemize}
\item Limiting the number of insights on similar topics.
\item Adjusting the weight of different pattern types based on user feedback.
\end{itemize}

\paragraph{interpretable}



Interpretability scoring represents the explainability of the model inputs and outputs, measuring the ability to justify why a certain data range is considered insightful.

For insights, interpretability is related to the significance of the insights. The more significant the insight, the higher its interpretability. Additionally, interpretability is associated with the prominence of pattern recognition.

\subsection{Context Storytelling}

Context-based data storytelling is one of the key components of the Multidimensional Data Storytelling Framework (MDSF). The context encompasses the user's editing history and modifications. By leveraging historical data and a large model Agent framework for reasoning, MDSF performs real-time analysis of the user's areas of interest, generating high-quality data reports and insightful analyses.

\subsubsection{Context Agent}



At the heart of the Big Model Agent framework is the use of pre-trained large language models (LLMS) such as GPT-4 for data analysis and text generation. The framework consists of the following key steps:

\begin{enumerate}
    \item Context understanding: The model first needs to understand the context of the user input, including the structure, dimensions, and metrics of the data.
          Using natural language processing techniques, the model is able to identify key information in the data and generate initial insights.
    \item Inference \& Analysis: After understanding the context, the model performs inference to identify patterns and anomalies in the data.
          This step often involves complex statistical analysis and machine learning algorithms such as time series analysis, clustering analysis, and regression analysis.
    \item Generate insights: Based on the inference results, the model generates a detailed insight report.
          These reports not only contain descriptive statistics of the data, but also provide predictive analysis and decision making recommendations.

\end{enumerate}

\subsubsection{Model Training for Enhanced Insight Scores}

\paragraph{Insigth Discovery \& Data Annotation}



Figure \ref{fig:framework} (a) illustrates the overall architecture of our proposed Multiscenario Data Storytelling Framework (MDSF), which includes both offline and online computation components. In the offline computation part, we first collect metadata from multiple data sources (such as databases and file systems) and conduct preliminary analysis through an automated data discovery process. This process employs various data mining and statistical analysis techniques to extract potential insights from the raw data. These insights are presented in various forms, including time series analysis, scatter plot analysis, and histograms, providing a foundation for subsequent insight computation.

After completing the preliminary automated insight discovery, we introduce human intervention to rank the insights. This manual ranking incorporates the knowledge and experience of domain experts, ensuring the high quality and relevance of the selected insights. The involvement of domain experts not only enhances the accuracy of the insights but also increases their practicality and interpretability. The output of this step is a preliminarily ranked list of insights, which serves as high-quality data input for subsequent model refinement.

\paragraph{Rank Model Training}




Next, we fine-tune the large language model (LLM) using the LLaMA-Factory framework to better understand and process domain-specific insights. Through this series of processes, we generate a fine-tuned model capable of accurately computing and ranking insights in subsequent online computations. The fine-tuned model not only improves the quality of generated insights but also enhances the model's adaptability to domain-specific content.

The online computation component primarily involves dynamically adjusting the ranking of insights based on the user's editing context and multi-scenario prompts. While the user is editing content, the MDSF framework parses the user's editing history, user profile, and current multi-scenario prompt information in real time. By comprehensively analyzing this information, the system can extract the most relevant insights from the precomputed insight library and re-rank these insights using the LLM. Specifically, the system first parses the data to identify potential insights in the current editing context. Then, it scores and ranks these insights using the LLM, generating descriptive text and visualizations. Finally, the system selects the optimal insights from the generated results, providing them to the user for further editing and utilization.

In this part, the combination of automation and human intervention ensures the high quality and relevance of the insights. Additionally, the dynamic adjustment mechanism in the online component allows the system to provide the most relevant insights in real time based on the user's editing context, thus enhancing the user's editing efficiency and content quality.

\section{Experiments}


In this section, we evaluate MDSF on four classes of tasks.
The specific research questions through the experimental study are as follows:

\begin{itemize}    
    \item[\textbf{RQ1}] Can MDSF understand insights and produce high-quality rankings?
    \item[\textbf{RQ2}] Can MDSF generate accurate and stable insight descriptions?
    \item[\textbf{RQ3}] Can MDSF create superior data stories in the data story generation task based on the above?
\end{itemize}

\subsection{Setup}



In this paper, we want to design a general framework that can be used on multiple models, so we refer to the latest open source and closed source models for the choice of large language model base.
We selected GPT-3-turbo, GPT4, Gemini-1.5, LLama3-7B, and Qwen2-7B as the base models.
Meanwhile, this paper also fine-tunes llama3-7B and qwen2-7B for the rank task and the generation task.

In the experiments, we use 2 NVIDIA V100 Gpus for fine-tuning, and the batch size of each GPU is 32.
We used the Adam optimizer with a learning rate of 1e-4 and trained for 10 epochs.

\subsubsection{Datasets}






We conducted experiments with the MDSF on one private dataset and three public datasets (Text2Analysis, InsightBench, and competition data) to validate its effectiveness and performance across different scenarios.

The private dataset is derived from the company's actual business data, encompassing information from multiple domains such as sales data, user behavior data, and advertisement data. Based on this dataset, we performed bulk data annotation to construct a benchmark dataset for multidimensional data analysis. Additionally, our analysts created multiple data reports and insight ranking samples from the insights contained within this dataset.

For the Text2Analysis dataset, it is a benchmark dataset specifically designed for table-based question answering, containing instances of advanced data analysis and ambiguous queries. We treated these complex queries as challenges for model inference, with positive examples consisting of correctly parsed and answered queries, and negative examples comprising queries that were either incorrectly parsed or inaccurately answered.

For the competition dataset, we utilized publicly available datasets from Kaggle, which cover information from various domains, including vehicle sales, Airbnb market analysis, real estate sales, and electric vehicle ownership. We selected multi-step reasoning tasks related to business decisions and labeled each instance as positive or negative based on whether the model could correctly predict and generate relevant insights according to the dataset's characteristics.

In the InsightBench dataset, we focused on evaluating MDSF's performance in multi-step insight generation tasks. This dataset is specifically designed to assess business analysis agents and includes insight generation tasks from multiple real business scenarios. The tasks require the model to not only understand complex business needs but also generate coherent and meaningful business insights through multi-step reasoning.

We partitioned these datasets finally. The oldest data instances were used to build the retrieval set, the newest instances formed the test set, and the intermediate instances were used for training. While we are aware of other potential data partitioning methods (e.g., stacking by time periods), due to space constraints, we adopted the aforementioned setup for this experiment and leave empirical studies on other configurations for future work.

\subsubsection{Evaluation Metrics}


We used multiple evaluation metrics to assess the performance of MDSF. For the ranking tasks, we employed the Spearman Footrule Distance.

\begin{equation}
    \text{F} = \sum_{i=1}^{n} |r_{i} - \hat{r}_{i}|
\end{equation}

$F$ is the Spearman Footrule, $r_{i}$ is the rank of the $i$-th item in the ground truth, and $\hat{r}_{i}$ is the rank of the $i$-th item in the prediction.
Bigger Spearman Footrule means worse performance.


For the insight description tasks, we used accuracy metrics, as we are more interested in whether the model can accurately represent the data without generating hallucinations.

In the accuracy evaluation, we focus on whether the generated descriptions accurately convey the type of insight, the content of the insight, and the value of the insight. Since this part of the evaluation is manually annotated, we conducted a sampled evaluation.


For the story generation task, we used BLEU and ROUGE as evaluation metrics.

\begin{equation}
    \text{BLEU} = \text{BP} \times \exp\left(\sum_{n=1}^{N} w_{n} \log p_{n}\right)
\end{equation}


BLEU and ROUGE are commonly used evaluation metrics in the field of NLP for assessing the quality of generated text. BLEU evaluates the quality of generated text by calculating the overlap of n-grams between the generated text and the reference text. ROUGE, on the other hand, assesses the quality of generated text by computing the overlap between the generated text and the reference text.

The choice of BLEU and ROUGE as evaluation metrics is due to their effectiveness in reflecting the degree of overlap between the generated text and the reference text, thereby providing a better assessment of the quality of the generated text.

\subsection{Experiment Results}


We evaluated the performance of MDSF on four types of tasks, 
and we present the experimental results of each task separately below.

\subsubsection{Insight Rank Task}

\begin{table}[h]
    \centering
    \begin{tabular}{ccc}
        \hline
        \textbf{Model} & \textbf{Private Dataset} & \textbf{InsightBench} \\ \hline
        Manual & 0.00 & 0.00 \\
        GPT-3.5 turbo & 8.35 & 12.10 \\
        GPT-4 & 5.83 & 7.00 \\
        Gemini 1 .5 & 7.21 & 8.25 \\ \hline
        Llama3-7B & 12.48 & 12.50 \\
        ChatGLM2-6B & 10.73 & 11.75 \\
        Qwen-2-7B-Chat & 9.17 & 9.50 \\
        Yi-1.5-9B & 11.25 & 12.00 \\ \hline
        MDSF & 6.82 & 7.25 \\ \hline
    \end{tabular}
    \caption{Performance comparison of different models based on Private Dataset and Insight bench SFD metric}
    \label{tab:model_sfd}
\end{table}


In the Insight Rank task, we compared the performance of the MDSF model with other mainstream models on both a private dataset and the InsightBench dataset. Table \ref{tab:model_sfd} presents the specific results of various models on the SFD metric, where a smaller SFD value indicates lower distance error and better performance.

As shown in the table, the manually annotated SFD value is 0.00, which serves as our benchmark. The GPT-4 model outperformed GPT-3.5 turbo on both datasets, achieving SFD values of 5.83 and 7.00, respectively. The Gemini 1.5 model also performed well, with SFD values of 7.21 on the private dataset and 8.25 on InsightBench.

In contrast, the Llama3-7B and ChatGLM2-6B models demonstrated relatively poorer performance, especially on the private dataset where Llama3-7B had an SFD value as high as 12.48. The Qwen-2-7B-Chat and Yi-1.5-9B models also underperformed, with SFD values exceeding 9.00.

Notably, the MDSF model achieved SFD values of 6.82 on the private dataset and 7.25 on InsightBench. Although it did not reach the level of manual annotation, it showed significant improvement over other automated models. This indicates that the MDSF model possesses good accuracy and reliability in handling the Insight Rank task.

In summary, the MDSF model outperformed most existing mainstream models in the Insight Rank task, demonstrating its advantage in controlling distance error.

\subsubsection{Insight Description Task}



In the Insight Description task, we used the ACC (accuracy) metric to evaluate the performance of different models. Figure \ref{fig:acc} shows the performance of each model on the ACC metric. A higher ACC value indicates greater accuracy of the model in the description task.
As shown in the figure \ref{fig:acc}, there are significant differences in the ACC performance of the various models:

\begin{figure}[h]
    \centering
    \includegraphics[width=1\linewidth]{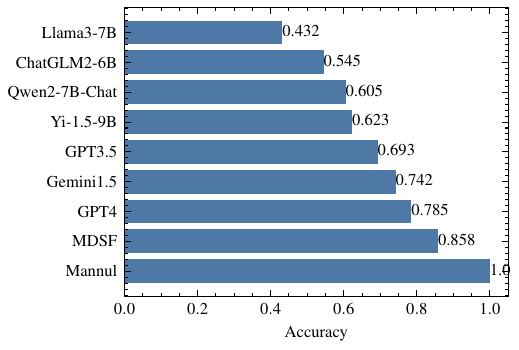}
    \caption{Performance comparison of different models based on ACC metric}
    \label{fig:acc}
\end{figure}



The specific results are as follows:

Llama3-7B had an ACC value of 0.432, indicating the poorest performance.
ChatGLM2-6B had an ACC value of 0.545, slightly higher than Llama3-7B.
Qwen2-7B-Chat and Yi-1.5-9B had ACC values of 0.605 and 0.623, respectively, representing moderate performance.
GPT-3.5 had an ACC value of 0.693, showing good performance.
Gemini 1.5 had an ACC value of 0.742, demonstrating excellent performance.
GPT-4 had an ACC value of 0.785, indicating outstanding performance.
The MDSF model had an ACC value of 0.858, the best among the automated models, second only to the manually annotated (Manual) value of 1.0.

From the results, it is evident that the MDSF model exhibited the most exceptional performance in the Insight Description task, achieving an ACC value of 0.858, significantly surpassing other automated models and approaching the accuracy of manual annotation. This indicates that the MDSF model has high accuracy and reliability in handling description tasks.

\subsubsection{Data Story Generation Task}

In the Data Story Generation task, we compared the performance of different models on a private dataset, Kaggle dataset, Text2Analysis dataset, and InsightBench dataset. Table \ref{tab:my-table} lists the scores of each model on the Rouge and BLEU metrics.

\begin{table*}[htp]
    \begin{center}
        \begin{tabular}{ccccccccc}
            \hline
            \multirow{2}{*}{\textbf{Model}} &
              \multicolumn{2}{c}{\textbf{PrivateDataset}} &
              \multicolumn{2}{c}{\textbf{Kaggle}} &
              \multicolumn{2}{c}{\textbf{Text2Analysis}} &
              \multicolumn{2}{c}{\textbf{InsightBench}} \\ \cline{2-9} &
              \textbf{Rouge} &
              \textbf{BLEU} &
              \textbf{Rouge} &
              \textbf{BLEU} &
              \textbf{Rouge} &
              \textbf{BLEU} &
              \textbf{Rouge} &
              \textbf{BLEU} \\ \hline
              GPT3.5-turbo  & 0.742 & 0.728 & 0.745 & 0.703 & 0.711 & 0.688 & 0.823 & 0.789 \\
              GPT4          & 0.894 & 0.853 & 0.865 & 0.827 & 0.842 & 0.745 & 0.952 & 0.915 \\
              Gemini1.5     & 0.876 & 0.849 & 0.871 & 0.832 & 0.847 & 0.712 & 0.937 & 0.922 \\ \hline
              Llama3-7B     & 0.673 & 0.591 & 0.689 & 0.650 & 0.652 & 0.589 & 0.714 & 0.705 \\
              Qwen2-7B      & 0.718 & 0.624 & 0.681 & 0.643 & 0.756 & 0.674 & 0.768 & 0.729 \\
              Llama3-7B-FT  & 0.726 & 0.619 & 0.669 & 0.630 & 0.805 & 0.712 & 0.759 & 0.719 \\
              Qwen2-7B-FT   & 0.747 & 0.674 & 0.665 & 0.627 & 0.813 & 0.708 & 0.754 & 0.716 \\ \hline
              \textbf{MDSF} & 0.862 & 0.825 & 0.780 & 0.736 & 0.752 & 0.813 & 0.864 & 0.824 \\ \hline
            \end{tabular}
            \caption{Performance comparison of different models based on Rouge and BLEU metrics}
            \label{tab:my-table}
    \end{center}

\end{table*}


From the table, it can be seen that GPT-4 and Gemini 1.5 outperformed other models on multiple datasets. Specifically, on the Kaggle dataset and InsightBench dataset, GPT-4 achieved Rouge and BLEU scores of 0.865 and 0.827, as well as 0.952 and 0.915, respectively. Gemini 1.5 also performed very close to these scores, with 0.871 and 0.832 on the Kaggle dataset, and 0.937 and 0.922 on InsightBench.

In contrast, Llama3-7B and Qwen2-7B performed relatively poorly, especially on the Kaggle dataset, where Llama3-7B had Rouge and BLEU scores of only 0.689 and 0.650. Even the fine-tuned versions, Llama3-7B-FT and Qwen2-7B-FT, did not significantly improve their performance.

Our proposed MDSF model exhibited stable performance across all datasets. Particularly on the Text2Analysis dataset and InsightBench dataset, the MDSF model achieved Rouge and BLEU scores of 0.752 and 0.813, and 0.864 and 0.824, respectively. While it did not surpass GPT-4 and Gemini 1.5, it still demonstrated strong competitiveness.

In summary, although GPT-4 and Gemini 1.5 showed outstanding performance in the Data Story Generation task, the MDSF model also performed well across multiple datasets, indicating its potential in data story generation tasks.





\subsubsection{User Study}


To comprehensively evaluate the performance of the MDSF model, we conducted a user survey. The evaluation criteria for the user survey included three aspects: Structure, Conclusion Extraction, and Richness. Table \ref{tab:my-table1} presents the specific evaluation criteria and scoring details.

\begin{table*}[]
    \centering
    \begin{tabular}{cccc}
    \hline
    \textbf{Evaluation Criteria} &
        \textbf{Structure} &
        \textbf{Conclusion Extraction} &
        \textbf{Richness of Detail} \\ \hline
    \textbf{0} &
        \begin{tabular}[c]{@{}c@{}}Chaotic structure\\ poor readability\end{tabular} &
        No conclusion extraction &
        No further elaboration \\ \hline
    \textbf{1} &
        \begin{tabular}[c]{@{}c@{}}Some analysis structure\\ but incomplete\end{tabular} &
        \begin{tabular}[c]{@{}c@{}}Module exists \\ but no valuable information\end{tabular} &
        Only a little elaboration \\ \hline
    \textbf{2} &
        \begin{tabular}[c]{@{}c@{}}Complete structure including \\ overview, details, suggestions\end{tabular} &
        \begin{tabular}[c]{@{}c@{}}Module exists \\ but little valuable information\end{tabular} &
        Elaboration without data support \\ \hline
    \textbf{3} &
        - &
        \begin{tabular}[c]{@{}c@{}}Module exists with \\ some valuable information\end{tabular} &
        \begin{tabular}[c]{@{}c@{}}Covers few dimensions \\ with data support\end{tabular} \\ \hline
    \textbf{4} &
        - &
        - &
        \begin{tabular}[c]{@{}c@{}}Covers almost all dimensions \\ with significant data support\end{tabular} \\ \hline
    \textbf{5} &
        - &
        - &
        \begin{tabular}[c]{@{}c@{}}Covers almost all dimensions, \\ uses various analytical methods  \\ and has significant data arguments\end{tabular} \\ \hline
    \end{tabular}
    \caption{User study evaluation criteria}
    \label{tab:my-table1}
\end{table*}

\begin{figure*}[!htp]
    \centering
    \begin{subfigure}[b]{0.3\textwidth}
        \centering
        \includegraphics[width=\textwidth]{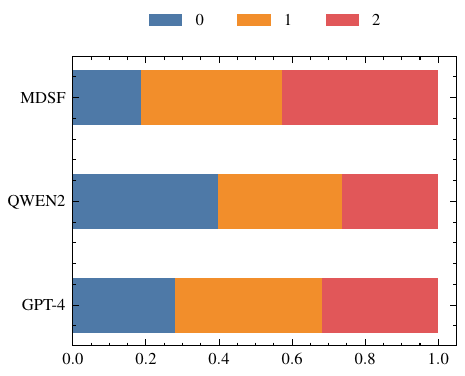}
        \caption{Structure}
        \label{fig:subfig1}
    \end{subfigure}
    \hfill
    \begin{subfigure}[b]{0.3\textwidth}
        \centering
        \includegraphics[width=\textwidth]{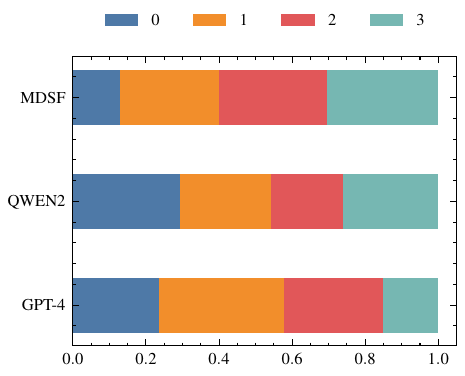}
        \caption{Conclusion Extraction}
        \label{fig:subfig2}
    \end{subfigure}
    \hfill
    \begin{subfigure}[b]{0.3\textwidth}
        \centering
        \includegraphics[width=\textwidth]{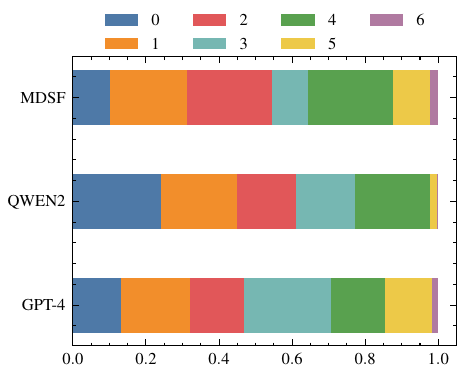}
        \caption{Richness of Detail}
        \label{fig:subfig3}
    \end{subfigure}
    \caption{User study results}
    \label{fig:main}
\end{figure*}

In the user survey, participants rated the outputs of each model based on the aforementioned criteria. Figure \ref{fig:main} presents the results of the user survey, showing the percentage scores of each model in terms of Structure, Conclusion Extraction, and Richness.

The observations from Figure \ref{fig:main} reveal the following key insights:

\begin{enumerate}
\item Structural Evaluation (Figure \ref{fig:subfig1}): The MDSF model demonstrates a significant advantage over Qwen2 and GPT-4 in structural integrity and readability. The ratings predominantly cluster around scores of 2 and 3, underscoring the model’s effectiveness in maintaining coherent and well-organized outputs.

\item Conclusion Extraction (Figure \ref{fig:subfig2}): The MDSF model performs notably well in extracting conclusions, with ratings concentrated around 2 and 3. This suggests a superior capability to distill and extract key insights and valuable information from the data.

\item Richness of Content (Figure \ref{fig:subfig3}): The MDSF model excels in providing rich, multi-dimensional content, as reflected in the distribution of ratings around 4 and 5. This demonstrates its ability to offer comprehensive analyses supported by diverse data dimensions.
\end{enumerate}

Overall, the user study findings indicate that the MDSF model achieves exceptional performance in structure, conclusion extraction, and content richness, substantially outperforming the comparative models. These results further establish the model’s reliability and practicality for augmented analysis in real-world scenarios.

\section{Conclusion}

This paper presents the Multi-dimensional Data Storytelling Framework (MDSF), which leverages large language models to automate the complexities of data analysis and context-aware report generation. Traditional data analysis systems often struggle to extract actionable insights and maintain consistency across diverse data sources. MDSF addresses these challenges by integrating automated data exploration, insight prioritization, and fine-tuned model continuations into a unified solution.

The framework effectively extracts and ranks actionable insights from large datasets, enhancing the analytical relevance and depth of generated reports. A pre-labeled fine-tuning phase ensures the large language model is both accurate and context-sensitive, enabling precise and meaningful outputs. The agent-based architecture facilitates seamless continuation control, even in scenarios with complex data histories and diverse contextual inputs, a critical capability for real-world applications characterized by data heterogeneity and contextual nuances. 

Experimental results on various real-world datasets demonstrate MDSF's superior accuracy, operational efficiency, and user satisfaction compared to existing methods. User feedback further underscores its practical value and potential for future refinement. 

In conclusion, MDSF integrates advanced language models with innovative insight management strategies, addressing key challenges in data analysis while setting a solid foundation for more sophisticated and user-focused data storytelling solutions.

\section*{Acknowledgment}

Thanks to the anonymous reviewers for improving this work.

\newpage

\bibliographystyle{IEEEtran}

\begin{thebibliography}{00}

\bibitem{1} M. S. Islam, M. T. R. Laskar, M. R. Parvez, E. Hoque, and S. Joty, DataNarrative: Automated Data-Driven Storytelling with Visualizations and Texts, Aug. 09, 2024. Accessed: Aug. 27, 2024. [Online]. Available: https://arxiv.org/abs/2408.05346v2
\bibitem{2} G. Sahu et al., InsightBench: Evaluating Business Analytics Agents Through Multi-Step Insight Generation, Jul. 08, 2024, arXiv: arXiv:2407.06423. Accessed: Jul. 18, 2024. [Online]. Available: http://arxiv.org/abs/2407.06423
\bibitem{3} Magic Quadrant for Analytics and Business Intelligence Platforms, Jun. 2024. Accessed: Jul. 10, 2024. [Online]. Available: https://www.gartner.com/doc/reprints?id=1-2HWZ69DK\&ct=240624\&st
\bibitem{4} Y. Xie, Y. Luo, G. Li, and N. Tang, HAIChart: Human and AI Paired Visualization System, Jun. 16, 2024, arXiv: arXiv:2406.11033. Accessed: Jul. 24, 2024. [Online]. Available: http://arxiv.org/abs/2406.11033
\bibitem{5} Y. Zhu, S. Du, B. Li, Y. Luo, and N. Tang, Are Large Language Models Good Statisticians?, Jun. 11, 2024, arXiv: arXiv:2406.07815. Accessed: Jul. 24, 2024. [Online]. Available: http://arxiv.org/abs/2406.07815
\bibitem{6} G. Li, R. Li, Y. Feng, Y. Zhang, Y. Luo, and C. H. Liu, CoInsight: Visual Storytelling for Hierarchical Tables With Connected Insights, IEEE Trans. Vis. Comput. Graphics, vol. 30, no. 6, pp. 3049–3061, Jun. 2024, doi: 10.1109/TVCG.2024.3388553.
\bibitem{7} S. Guo, C. Deng, Y. Wen, H. Chen, Y. Chang, and J. Wang, DS-Agent: Automated Data Science by Empowering Large Language Models with Case-Based Reasoning, May 28, 2024, arXiv: arXiv:2402.17453. Accessed: Jun. 05, 2024. [Online]. Available: http://arxiv.org/abs/2402.17453
\bibitem{8} H. Shao, R. Martinez-Maldonado, V. Echeverria, L. Yan, and D. Gasevic, Data Storytelling in Data Visualisation: Does it Enhance the Efficiency and Effectiveness of Information Retrieval and Insights Comprehension?, in Proceedings of the CHI Conference on Human Factors in Computing Systems, May 2024, pp. 1–21. doi: 10.1145/3613904.3643022.
\bibitem{9} A. Singha, B. Chopra, A. Khatry, S. Gulwani, and A. Z. Henley, Semantically Aligned Question and Code Generation for Automated Insight Generation, Apr. 2024.
\bibitem{10} X. Lan, L. Yang, Z. Wang, Y. Wang, D. Shi, and S. Carpendale, Gen4DS: Workshop on Data Storytelling in an Era of Generative AI, Apr. 05, 2024, arXiv: arXiv:2404.01622. Accessed: Jul. 11, 2024. [Online]. Available: http://arxiv.org/abs/2404.01622
\bibitem{12} Y. Sui, M. Zhou, M. Zhou, S. Han, and D. Zhang, Table Meets LLM: Can Large Language Models Understand Structured Table Data? A Benchmark and Empirical Study, in Proceedings of the 17th ACM International Conference on Web Search and Data Mining, Merida Mexico: ACM, Mar. 2024, pp. 645–654. doi: 10.1145/3616855.3635752.
\bibitem{13} Y. He, S. Cao, Y. Shi, Q. Chen, K. Xu, and N. Cao, Leveraging Large Models for Crafting Narrative Visualization: A Survey, Jan. 25, 2024, arXiv: arXiv:2401.14010. Accessed: Apr. 28, 2024. [Online]. Available: http://arxiv.org/abs/2401.14010
\bibitem{14} L. Shen, H. Li, Y. Wang, and H. Qu, From Data to Story: Towards Automatic Animated Data Video Creation with LLM-based Multi-Agent Systems, 2024, arXiv: arXiv:2408.03876. doi: 10.48550/arXiv.2408.03876.
\bibitem{16} X. He et al., Text2Analysis: A Benchmark of Table Question Answering with Advanced Data Analysis and Unclear Queries, Dec. 21, 2023, arXiv: arXiv:2312.13671. Accessed: Feb. 27, 2024. [Online]. Available: http://arxiv.org/abs/2312.13671
\bibitem{17} R. Ding, S. Han, and D. Zhang, InsightPilot: An LLM-Empowered Automated Data Exploration System, in EMNLP 2023, ACL special interest group on linguistic data (SIGDAT), Dec. 2023. [Online]. Available: https://www.microsoft.com/en-us/research/publication/insightpilot-an-llm-empowered-automated-data-exploration-system/
\bibitem{18} Y. Zhao, H. Zhang, S. Si, L. Nan, X. Tang, and A. Cohan, Investigating Table-to-Text Generation Capabilities of LLMs in Real-World Information Seeking Scenarios, Oct. 30, 2023, arXiv: arXiv:2305.14987. Accessed: Jun. 05, 2024. [Online]. Available: http://arxiv.org/abs/2305.14987
\bibitem{19} H. Li, J. Su, Y. Chen, Q. Li, and Z. Zhang, SheetCopilot: Bringing Software Productivity to the Next Level through Large Language Models, Oct. 2023. doi: 10.48550/arXiv.2305.19308.
\bibitem{21} F. Stalph and B. Heravi, Exploring Data Visualisations: An Analytical Framework Based on Dimensional Components of Data Artefacts in Journalism, Digital Journalism, vol. 11, no. 9, pp. 1641–1663, Oct. 2023, doi: 10.1080/21670811.2021.1957965.
\bibitem{22} P. Li et al., Table-GPT: Table-tuned GPT for Diverse Table Tasks, Oct. 13, 2023, arXiv: arXiv:2310.09263. Accessed: Jun. 04, 2024. [Online]. Available: http://arxiv.org/abs/2310.09263
\bibitem{24} L. Wang, S. Zhang, Y. Wang, E.-P. Lim, and Y. Wang, LLM4Vis: Explainable Visualization Recommendation using ChatGPT, Oct. 2023, Accessed: May 07, 2024. [Online]. Available: https://www.microsoft.com/en-us/research/publication/llm4vis-explainable-visualization-recommendation-using-chatgpt/
\bibitem{25} H. Li, Y. Wang, and H. Qu, Where Are We So Far? Understanding Data Storytelling Tools from the Perspective of Human-AI Collaboration, Sep. 27, 2023. Accessed: Aug. 27, 2024. [Online]. Available: https://arxiv.org/abs/2309.15723v2
\bibitem{26} C. Wang, J. Thompson, and B. Lee, Data Formulator: AI-powered Concept-driven Visualization Authoring, Sep. 18, 2023. Accessed: Nov. 22, 2023. [Online]. Available: https://arxiv.org/abs/2309.10094v2
\bibitem{27} G. Renda, M. Daquino, and V. Presutti, Melody: A Platform for Linked Open Data Visualisation and Curated Storytelling, 34TH ACM CONFERENCE ON HYPERTEXT AND SOCIAL MEDIA, HT 2023. ASSOC COMPUTING MACHINERY, 1601 Broadway, 10th Floor, NEW YORK, NY, UNITED STATES, pp. 1–8, Sep. 04, 2023. doi: 10.1145/3603163.3609035.
\bibitem{28} P. Wang et al., Large Language Models are not Fair Evaluators, Aug. 30, 2023, arXiv: arXiv:2305.17926. Accessed: Jul. 17, 2024. [Online]. Available: http://arxiv.org/abs/2305.17926
\bibitem{30} L. Zheng, N. Li, X. Chen, Q. Gan, and W. Zhang, Dense Representation Learning and Retrieval for Tabular Data Prediction, in Proceedings of the 29th ACM SIGKDD Conference on Knowledge Discovery and Data Mining, Long Beach CA USA: ACM, Aug. 2023, pp. 3559–3569. doi: 10.1145/3580305.3599305.
\bibitem{31} S. Mysore, A. McCallum, and H. Zamani, Large Language Model Augmented Narrative Driven Recommendations, Jul. 21, 2023, arXiv: arXiv:2306.02250. Accessed: Dec. 06, 2023. [Online]. Available: http://arxiv.org/abs/2306.02250
\bibitem{32} Z. Zhao and N. Elmqvist, The Stories We Tell About Data: Surveying Data-Driven Storytelling Using Visualization, IEEE Comput. Graph. Appl., vol. 43, no. 4, pp. 97–110, Jul. 2023, doi: 10.1109/MCG.2023.3269850.
\bibitem{33} P. Yin et al., Natural Language to Code Generation in Interactive Data Science Notebooks, in Proceedings of the 61st Annual Meeting of the Association for Computational Linguistics (Volume 1: Long Papers), A. Rogers, J. Boyd-Graber, and N. Okazaki, Eds., Toronto, Canada: Association for Computational Linguistics, Jul. 2023, pp. 126–173. doi: 10.18653/v1/2023.acl-long.9.
\bibitem{34} V. Dibia, LIDA: A Tool for Automatic Generation of Grammar-Agnostic Visualizations and Infographics using Large Language Models, in Proceedings of the 61st Annual Meeting of the Association for Computational Linguistics (Volume 3: System Demonstrations), D. Bollegala, R. Huang, and A. Ritter, Eds., Toronto, Canada: Association for Computational Linguistics, Jul. 2023, pp. 113–126. doi: 10.18653/v1/2023.acl-demo.11.
\bibitem{35} W. Zhang, Y. Shen, W. Lu, and Y. Zhuang, Data-Copilot: Bridging Billions of Data and Humans with Autonomous Workflow, Jun. 12, 2023, arXiv: arXiv:2306.07209. doi: 10.48550/arXiv.2306.07209.
\bibitem{36} X. Wu et al., Holistic Cube Analysis: A Query Framework for Data Insights, Jun. 04, 2023, arXiv: arXiv:2302.00120. doi: 10.48550/arXiv.2302.00120.
\bibitem{37} C. Chai, N. Tang, J. Fan, and Y. Luo, Demystifying Artificial Intelligence for Data Preparation, in Companion of the 2023 International Conference on Management of Data, Seattle WA USA: ACM, Jun. 2023, pp. 13–20. doi: 10.1145/3555041.3589406.
\bibitem{39} L. Shen et al., Towards Natural Language Interfaces for Data Visualization: A Survey, IEEE Trans. Visual. Comput. Graphics, vol. 29, no. 6, pp. 3121–3144, Jun. 2023, doi: 10.1109/TVCG.2022.3148007.
\bibitem{40} P. Ma, R. Ding, S. Wang, S. Han, and D. Zhang, XInsight: eXplainable Data Analysis Through The Lens of Causality, May 30, 2023, arXiv: arXiv:2207.12718. Accessed: Jun. 19, 2023. [Online]. Available: http://arxiv.org/abs/2207.12718
\bibitem{41} P. Maddigan and T. Susnjak, Chat2VIS: Generating Data Visualizations via Natural Language Using ChatGPT, Codex and GPT-3 Large Language Models, IEEE Access, vol. 11, pp. 45181–45193, May 2023, doi: 10.1109/ACCESS.2023.3274199.
\bibitem{42} C. Harris et al., SpotLight: Visual Insight Recommendation, in Companion Proceedings of the ACM Web Conference 2023, Austin TX USA: ACM, Apr. 2023, pp. 19–23. doi: 10.1145/3543873.3587302.
\bibitem{43} W. Li et al., NetworkNarratives: Data Tours for Visual Network Exploration and Analysis, in Proceedings of the 2023 CHI Conference on Human Factors in Computing Systems, Apr. 2023, pp. 1–15. doi: 10.1145/3544548.3581452.
\bibitem{49} J. Zhao et al., ChartStory: Automated Partitioning, Layout, and Captioning of Charts into Comic-Style Narratives, IEEE Trans. Visual. Comput. Graphics, vol. 29, no. 2, pp. 1384–1399, Feb. 2023, doi: 10.1109/TVCG.2021.3114211.
\bibitem{50} P. S. Bobkowski and C. E. Etheridge, Spreadsheets, Software, Storytelling, Visualization, Lifelong Learning: Essential Data Skills for Journalism and Strategic Communication Students, Science Communication, vol. 45, no. 1, pp. 95–116, Feb. 2023, doi: 10.1177/10755470221147887.
\bibitem{51} S. Xu, E. Koh, F. Du, T. Y. Lee, S. M. Lee, and R. Rossi, Generating visual data stories, Jan. 24, 2023
\bibitem{53} D. Gkitsakis, S. Kaloudis, E. Mouselli, V. Peralta, P. Marcel, and P. Vassiliadis, Assessment Methods for the Interestingness of Cube Queries, 2023.
\bibitem{56} H. W. Chung et al., Scaling Instruction-Finetuned Language Models, Dec. 06, 2022, arXiv: arXiv:2210.11416. Accessed: Jul. 17, 2024. [Online]. Available: http://arxiv.org/abs/2210.11416
\bibitem{57} Q. Wang, Z. Chen, Y. Wang, and H. Qu, A Survey on ML4VIS: Applying Machine Learning Advances to Data Visualization, IEEE Trans Vis Comput Graph, vol. 28, no. 12, pp. 5134–5153, Dec. 2022, doi: 10.1109/TVCG.2021.3106142.
\bibitem{58} L. Shen, E. Shen, Z. Tai, Y. Xu, and J. Wang, Visual Data Analysis with Task-based Recommendations, Data Sci. Eng., vol. 7, no. 4, pp. 354–369, Dec. 2022, doi: 10.1007/s41019-022-00195-3.
\bibitem{59} Y. Zhang, M. Reynolds, A. Lugmayr, K. Damjanov, and G. M. Hassan, A Visual Data Storytelling Framework, Informatics, vol. 9, no. 4, p. 73, Sep. 2022, doi: 10.3390/informatics9040073.
\bibitem{60} D. Deng, A. Wu, H. Qu, and Y. Wu, DashBot: Insight-Driven Dashboard Generation Based on Deep Reinforcement Learning, Sep. 13, 2022, arXiv: arXiv:2208.01232. doi: 10.48550/arXiv.2208.01232.
\bibitem{63} V. Porwal et al., Efficient Insights Discovery through Conditional Generative Model based Query Approximation, in Proceedings of the 2022 International Conference on Management of Data, Philadelphia PA USA: ACM, Jun. 2022, pp. 2397–2400. doi: 10.1145/3514221.3520161.
\bibitem{65} E. Hosseini-Asl, W. Liu, and C. Xiong, A Generative Language Model for Few-shot Aspect-Based Sentiment Analysis, Apr. 11, 2022, arXiv: arXiv:2204.05356. Accessed: Nov. 22, 2023. [Online]. Available: http://arxiv.org/abs/2204.05356
\bibitem{67} D. Zdanovic, T. J. Lembcke, and T. Bogers, The Influence of Data Storytelling on the Ability to Recall Information, in Proceedings of the 2022 Conference on Human Information Interaction and Retrieval, in CHIIR ’22. New York, NY, USA: Association for Computing Machinery, Mar. 2022, pp. 67–77. doi: 10.1145/3498366.3505755.
\bibitem{70} Y. Zhou, X. Meng, Y. Wu, T. Tang, Y. Wang, and Y. Wu, An intelligent approach to automatically discovering visual insights, Journal of Visualization, pp. 1–18, 2022.
\bibitem{75} S. A. Matei and L. Hunter, Data storytelling is not storytelling with data: A framework for storytelling in science communication and data journalism, The Information Society, vol. 37, no. 5, pp. 312–322, Oct. 2021, doi: 10.1080/01972243.2021.1951415.
\bibitem{77} C. Hsu, Y.-W. Chu, T.-H. Huang, and L.-W. Ku, Plot and Rework: Modeling Storylines for Visual Storytelling, in Findings of the Association for Computational Linguistics: ACL-IJCNLP 2021, C. Zong, F. Xia, W. Li, and R. Navigli, Eds., Online: Association for Computational Linguistics, Aug. 2021, pp. 4443–4453. doi: 10.18653/v1/2021.findings-acl.390.
\bibitem{78} A. Wu et al., MultiVision: Designing Analytical Dashboards with Deep Learning Based Recommendation, Jul. 16, 2021, arXiv: arXiv:2107.07823. Accessed: Apr. 22, 2024. [Online]. Available: http://arxiv.org/abs/2107.07823
\bibitem{80} A. Personnaz, S. Amer-Yahia, L. Berti-Equille, M. Fabricius, and S. Subramanian, Balancing Familiarity and Curiosity in Data Exploration with Deep Reinforcement Learning, in Fourth Workshop in Exploiting AI Techniques for Data Management, in aiDM ’21. New York, NY, USA: Association for Computing Machinery, Jun. 2021, pp. 16–23. doi: 10.1145/3464509.3464884.
\bibitem{81} P. Ma, R. Ding, S. Han, and D. Zhang, MetaInsight: Automatic Discovery of Structured Knowledge for Exploratory Data Analysis, in Proceedings of the 2021 International Conference on Management of Data, Virtual Event China: ACM, Jun. 2021, pp. 1262–1274. doi: 10.1145/3448016.3457267.
\bibitem{83} N. Sultanum, F. Chevalier, Z. Bylinskii, and Z. Liu, Leveraging Text-Chart Links to Support Authoring of Data-Driven Articles with VizFlow, in Proceedings of the 2021 CHI Conference on Human Factors in Computing Systems, in CHI ’21. New York, NY, USA: Association for Computing Machinery, May 2021, pp. 1–17. doi: 10.1145/3411764.3445354.
\bibitem{84} A. Srinivasan, N. Nyapathy, B. Lee, S. M. Drucker, and J. Stasko, Collecting and Characterizing Natural Language Utterances for Specifying Data Visualizations, in Proceedings of the 2021 CHI Conference on Human Factors in Computing Systems, in CHI ’21. New York, NY, USA: Association for Computing Machinery, May 2021, pp. 1–10. doi: 10.1145/3411764.3445400.
\bibitem{87} D. Shi, X. Xu, F. Sun, Y. Shi, and N. Cao, Calliope: Automatic Visual Data Story Generation from a Spreadsheet, Ieee T Vis Comput Gr, vol. 27, no. 2, pp. 453–463, Feb. 2021, doi: 10.1109/TVCG.2020.3030403.
\bibitem{88} A. Wu et al., Ai4vis: Survey on artificial intelligence approaches for data visualization, IEEE Transactions on Visualization and Computer Graphics, 2021.
\bibitem{91} Q. Li et al., Exploring the" Double-Edged Sword" Effect of Auto-Insight Recommendation in Exploratory Data Analysis., in IUI Workshops, 2021.
\bibitem{92} A. Aghajanyan et al., HTLM: Hyper-Text Pre-Training and Prompting of Language Models, 2021, arXiv: arXiv:2107.06955. doi: 10.48550/arXiv.2107.06955.
\bibitem{93} T. B. Brown et al., Language Models are Few-Shot Learners, Jul. 22, 2020, arXiv: arXiv:2005.14165. Accessed: Dec. 27, 2022. [Online]. Available: http://arxiv.org/abs/2005.14165
\bibitem{94} W. Chen, J. Chen, Y. Su, Z. Chen, and W. Y. Wang, Logical Natural Language Generation from Open-Domain Tables, in Proceedings of the 58th Annual Meeting of the Association for Computational Linguistics, D. Jurafsky, J. Chai, N. Schluter, and J. Tetreault, Eds., Online: Association for Computational Linguistics, Jul. 2020, pp. 7929–7942. doi: 10.18653/v1/2020.acl-main.708.
\bibitem{95} T. Milo and A. Somech, Automating Exploratory Data Analysis via Machine Learning: An Overview, in Proceedings of the 2020 ACM SIGMOD International Conference on Management of Data, Portland OR USA: ACM, Jun. 2020, pp. 2617–2622. doi: 10.1145/3318464.3383126.
\bibitem{98} Y. Wang et al., DataShot: Automatic Generation of Fact Sheets from Tabular Data, IEEE Transactions on Visualization and Computer Graphics, vol. 26, no. 1, pp. 895–905, Jan. 2020, doi: 10.1109/TVCG.2019.2934398.
\bibitem{99} X. Qin, Y. Luo, N. Tang, and G. Li, Making data visualization more efficient and effective: a survey, The VLDB Journal, vol. 29, no. 1, pp. 93–117, Jan. 2020, doi: 10.1007/s00778-019-00588-3.
\bibitem{103} R. Ding, S. Han, Y. Xu, H. Zhang, and D. Zhang, QuickInsights: Quick and Automatic Discovery of Insights from Multi-Dimensional Data, in Proceedings of the 2019 International Conference on Management of Data, Amsterdam Netherlands: ACM, Jun. 2019, pp. 317–332. doi: 10.1145/3299869.3314037.

\end{thebibliography}

\end{document}